# Intelligent Traffic Monitoring with YOLOv11: A Case Study in Real-Time Vehicle Detection

Shkëlqim Sherifi
*Dept. of Computer Science*
*University of Tetovo*
Tetove, R.N.Macedonia
0009-0006-2227-5533

Festim Halili
*Dept. of Computer Science*
*University of Tetovo*
Tetove, R.N.Macedonia
0000-0001-7043-2173

Merita Kasa-Halili
*Dept. of Computer Science*
*University of Tetovo*
Tetove, R.N.Macedonia
0009-0001-9778-8820

*Abstract*— Recent advancements in computer vision, driven by artificial intelligence, have significantly enhanced monitoring systems. One notable application is traffic monitoring, which leverages computer vision alongside deep learning-based object detection and tracking models. These approaches demonstrate superior performance in detecting and counting vehicles, particularly cars, in traffic environments. This paper introduces an AI-powered traffic monitoring system designed to advance the development of a robust and efficient traffic surveillance architecture. The system integrates a pre-trained computer vision model YOLOv11 with the OpenCV library, the BoT-SORT and ByteTrack tracking algorithm, and the PyTorch framework to construct a convolutional neural network (CNN) that enhances the speed and efficiency of real-time vehicle detection from video frames. Additionally, the Qt library for Python is used to develop a desktop application that offers an intuitive and user-friendly interface with a clean design. The application supports offline usage on a local machine, ensuring accessibility and reliability without dependence on cloud infrastructure. Results of the experiments indicate that the efficiency of traffic monitoring can be significantly improved, achieving an accuracy of 66.67% - 95.83% in vehicle detection and counting. Results show that YOLOv11 achieves high precision (0.97-1.00 for cars, 1.00 for trucks), indicating effective minimization of false positives. Recall ranges from (0.82-1.00 for cars and 0.70-1.00 for trucks), suggesting better coverage for cars due to more training samples. F1-scores confirm robust performance for cars (0.90-1.00) and slightly more variable results for trucks (0.82-1.00), likely influenced by lighting and occlusion; nonetheless, the model attained perfect scores. However, adverse weather conditions may negatively impact this performance. Overall, this paper contributes to the modernization and development of future smart cities by showing the capacity of AI-driven traffic monitoring systems.

## I. Introduction

Artificial Intelligence (AI) is a human-engineered construct rather than a naturally occurring phenomenon; it comprises artifacts that, through complex computational processes, exhibit characteristics typically associated with intelligent behavior or traits not commonly attributed to artificial systems [10]. Webster's New World Dictionary defines "intelligence" as: "the ability to learn or understand or to deal with new or trying situations; the skilled use of reason; mental acuteness" [11].

Traffic monitoring has evolved from traditional, human-based, and inefficient surveillance methods to more modern, sensor-based digital approaches. In recent years, the rapid advancement of artificial intelligence, particularly through neural networks and computer vision, has further contributed to the development of more effective and intelligent traffic monitoring systems.

The findings presented in this paper are of considerable significance, as they introduce an innovative, AI-based solution for traffic monitoring. By automating and digitalizing the monitoring process, these systems substantially enhance operational reliability and efficiency.

The proposed system utilizes the YOLOv11 model to perform real-time vehicle detection on highway roads, with tracking facilitated by the BoT-SORT and ByteTrack algorithms. The model is trained on the COCO dataset [8]. The results are noteworthy, achieving frame rates of up to 290 FPS and approximately 76.8% mAP@50 on the COCO benchmark [20]. These results align with the state-of-the-art in real-time vehicle detection and tracking within the field of computer vision.

Recent advancements in computer vision and deep learning have enabled the development of highly efficient traffic monitoring systems. Building on this foundation, the present study formulates the following hypotheses:

**H1**: AI-based traffic monitoring systems do not meet the required speed and accuracy for real-world deployment.

**H2**: Adverse weather conditions significantly impair the functionality of AI-based traffic monitoring systems.

To validate the proposed hypotheses, we developed a real-time traffic monitoring system based on the YOLOv11 architecture. The system achieved detection accuracy ranging from 66.67% to 95.83%. During the initial processing phase, the system operated at 3.5 to 5.5 times slower than real-time video playback, due to the computational demands of frame-by-frame analysis. However, once a cached version of the processed video was generated, the system achieved real-time performance across all tested sequences. The model demonstrated high precision (0.97-1.00 for cars, 1.00 for trucks), indicating a low false positive rate. Recall ranged from 0.82-1.00 for cars and 0.70-1.00 for trucks, suggesting stronger detection performance for cars, likely due to a larger training dataset and more consistent visual features. The F1-score further confirmed this trend, with values between 0.90-1.00 for cars and 0.82-1.00 for trucks, reflecting the model's robustness in car detection. All experiments were conducted in a local environment using a GTX 960M 4GB GPU. These results directly refute H1, demonstrating that AI-based traffic monitoring systems can operate in real time with high accuracy. Furthermore, while H2 is partially supported since environmental factors such as lighting and weather conditions can affect detection accuracy, the system remained operational. This resilience is attributed to the robustness of the YOLOv11 model and the optimization of the tracking algorithms for challenging conditions.

This paper addresses the following research questions:

**RQ1**: How can an AI-based traffic monitoring system be implemented to achieve real-time performance with acceptable accuracy?

**RQ2**: In what ways can Artificial Intelligence improve the effectiveness of traffic monitoring?

**RQ3**: Are such systems scalable and financially viable for broader deployment?

The remainder of this paper is organized as follows: Section 1 provides a brief overview of Artificial Intelligence and Computer Vision technologies, defines the research questions addressed in this study, explains the methodology, describes the dataset used, outlines the tracking algorithms, presents a concise summary of the YOLOv11 architecture, and introduces the overall structure of the paper. Section 2 presents a concise review of the relevant literature. Section 3 discusses Computer Vision and image-based object recognition technologies. Section 4 describes the Vehicle Verification Process. Section 5 explains the operation of the AI and Computer Vision-based traffic monitoring system AIvision. Section 6 presents the performance metrics of the developed AIvision system. Finally, Section 7 concludes the paper by summarizing the main contributions, and Section 8 discusses the findings of this paper, outlines its limitations, and proposes directions for future research and development.

### A. Methodology

This study addresses the challenge of autonomous traffic monitoring through the application of Artificial Intelligence (AI), with a focus on real-time vehicle detection, tracking, and counting. A quantitative research methodology was adopted to evaluate the system's performance under varying environmental and operational conditions.

The proposed system, named AIvision, integrates several advanced technologies, including the YOLOv11 object detection model, the PyTorch deep learning framework, and the BoT-SORT and ByteTrack tracking algorithms. These components were selected for their proven efficiency in real-time computer vision tasks.

To assess the system's effectiveness, five distinct traffic video sequences captured at different locations and under varying lighting and weather conditions were analyzed. The evaluation focused on two key performance metrics: detection accuracy and processing speed. The results were quantitatively measured and presented in both tabular and graphical formats.

The research methodology followed a two-phase approach:

- *Exploratory Phase*: General information was gathered through a comprehensive review of existing literature, including peer-reviewed journals, conference proceedings, and reputable online sources.
- *Experimental Phase*: A focused investigation was conducted through the implementation and testing of the AIvision system. This phase involved empirical data collection from video analysis, enabling a detailed performance evaluation.

This methodological framework ensured both theoretical grounding and empirical validation, supporting the development of a robust and scalable AI-based traffic monitoring solution.

### B. Dataset

For the evaluation of YOLOv11 in vehicle detection, we utilized the Common Objects in Context (COCO) dataset, developed and maintained by Microsoft [25]. This dataset is widely recognized in the computer vision community for its diversity, scale, and high-quality annotations, making it a standard benchmark for object detection tasks.

The COCO dataset includes 330,000 images with over 200,000 labeled images, with more than 1.5 million object instances across 80 object categories [8]. For our specific use case, we focused on a subset of the dataset that includes vehicle-related classes such as cars, trucks, buses, motorcycles, and bicycles. Each image is annotated with precise bounding boxes and class labels, enabling supervised training of object detection models.

To ensure consistency and compatibility with the YOLOv11 architecture, all images are with a resolution of 640 × 640 pixels. The dataset includes a wide range of real-world conditions, including daytime and nighttime scenes, varying weather conditions (e.g., rain, fog, and shadows), and complex scenarios involving occlusions and overlapping vehicles.

The dataset was partitioned into 118,000 for training, 5,000 for validation, and 20,000 for testing, ensuring a balanced distribution for robust model evaluation.

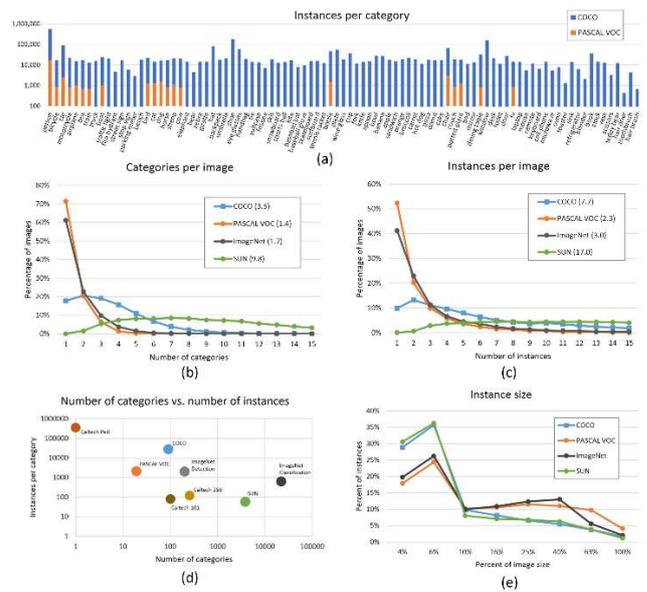

Fig. 1 (a) nr. of instances per category for MS COCO and PASCAL VOC. (b,c) nr. of categories and annotated instances for MS COCO, ImageNet Detection, PASCAL VOC, and SUN. (d) Number of categories vs. the number of instances per category. (e) The distribution of instance sizes for the MS COCO, ImageNet Detection, PASCAL VOC, and SUN datasets [25].

### C. Tracking Algorithms

All experiments related to the tracking algorithms were conducted using an Nvidia V100 GPU within the Google Colab environment.

TABLE I. TRACKING ALGORITHMS COMPARISON

| Tracker | Parameters | | | |
|---|---|---|---|---|
| | *Avr. Min FPS* | *Avr. FPS* | *Avr. Max FPS* | *Range of FPS* |
| BoTSort | 26.3 | 36.4 | 42.5 | 18.2 |
| ByteTrack | 40.6 | 57.3 | 69.3 | 26.8 |

To compute the Multiple Object Tracking Accuracy (MOTA), Equation (1) is used, where *i* denotes the frame index. The results of this computation are presented in Table II. For training, frames from the video file test1.mp4 were utilized.

Fig. 2 compares the frame rate performance of two tracking algorithms, BoTSort and ByteTrack, across four metrics: average minimum FPS, average FPS, average maximum FPS, and FPS range. ByteTrack achieves the highest minimum and maximum average FPS (40.6 and 69.3), while BoTSort records a higher minimum FPS (26.3). ByteTrack also shows a broader FPS range (26.8) compared to BoTSort (18.2), indicating greater variability. These results suggest that ByteTrack offers higher peak performance, whereas BoTSort provides more consistent frame rates.

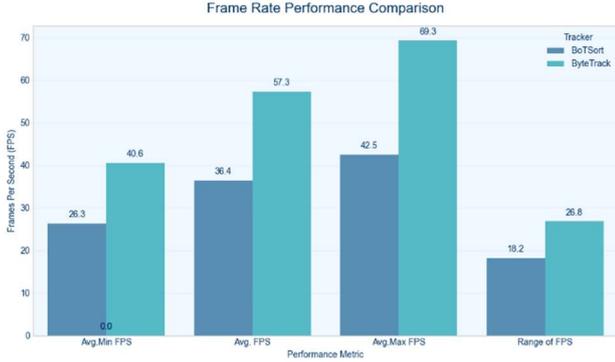

Fig. 2  BoTSort vs ByteTrack FPS Comparison

TABLE II.  MOTA RESULTS

| Tracker | Parameters | | | | |
|---|---|---|---|---|---|
| | GT | TP | FN | FP | IDS | MOTA |
| BoTSort | 61 | 64 | 0 | 3 | 1 | 0.9375 |
| ByteTrack | 61 | 63 | 0 | 2 | 0 | 0.968 |

$$MOTA = 1 - ((\sum i(FN_i + FP_i + IDS_i))/(\sum iGT_i)) \quad (1)$$

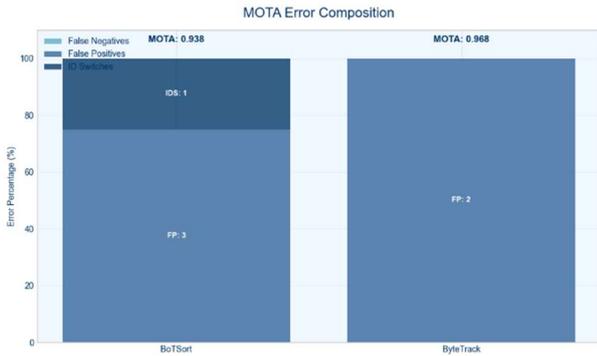

Fig. 3  BoTSort vs ByteTrack MOTA Comparison

Fig. 3 compares the error composition of two tracking algorithms, BoTSort and ByteTrack, based on the Multiple Object Tracking Accuracy (MOTA) metric. ByteTrack achieves a higher MOTA score of 0.968, with only 2% false positives and no ID switches. BoTSort follows with a MOTA of 0.938, comprising 3% false positives and 1% ID switches. These results indicate that ByteTrack provides slightly more accurate and stable tracking performance, with fewer identity errors and lower overall detection mistakes.

TABLE III.  MOTP RESULTS

| Tracker | Parameters | | |
|---|---|---|---|
| | Total IoU | Total Matches | MOTP |
| BoTSort | 61 | 64 | 0.953 |
| ByteTrack | 61 | 63 | 0.968 |

$$MOTP = 1 - ((\sum(t, i(dt, i)))/(\sum tCt)) \quad (2)$$

where $c_t$ is the number of matches in frame *t*, and $d_{t,\,i}$ is the *IoU* of target *i* with its corresponding *GT* label [26].

Fig. 4 compares the Multiple Object Tracking Precision (MOTP) scores of two tracking algorithms: BoTSort and ByteTrack. ByteTrack achieves a higher MOTP score of 0.968, indicating more precise localization of tracked objects. BoTSort follows with a slightly lower score of 0.953. These results suggest that ByteTrack offers better alignment between predicted and ground truth object positions.

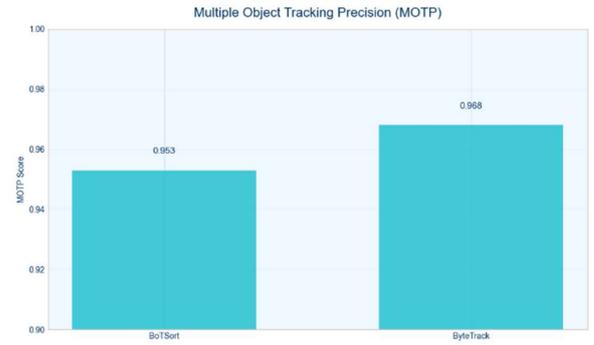

Fig. 4  BoTSort vs ByteTrack MOTP Comparison

D. YOLOv11 Architecture

The YOLOv11 architecture introduces several enhancements over previous versions by incorporating new layers, structural blocks, and optimization techniques aimed at improving both efficiency and detection accuracy [20].

Fig. 5 illustrates the structural composition of the YOLOv11 architecture.

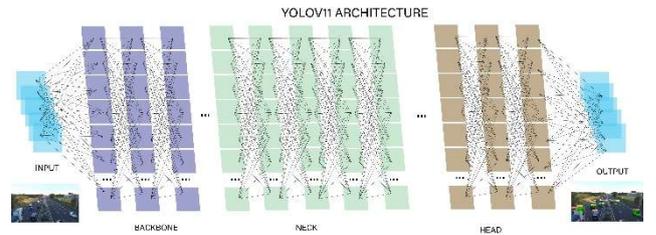

Fig. 5  YOLOv11 Architecture

The architecture is organized into three primary components:

*1) Backbone* is responsible for multi-scale feature extraction from input images. It includes [20]:
- *Convolutional Layers* for progressive down-sampling and depth enhancement.

$$Conv1 = Conv(I, 64, 3, 2) \quad (3)$$

$$Conv2 = Conv(Conv1, 128, 3, 2) \quad (4)$$

- *C3k2 Block*, a more efficient alternative to the C2f block, based on Cross-Stage Partial (CSP) networks, using smaller kernel convolutions to reduce computational cost.

$$C3k2(X) = Conv(Split(X))$$
$$+ Conv\big(Merge(Split(X))\big) \quad (5)$$

- *SPPF Block (Spatial Pyramid Pooling Fast)*, retained from earlier versions, performs multi-scale spatial pooling.

$$SPPF(X) = Concat(MaxPool(X, 5),$$
$$MaxPool(X, 3), MaxPool(X, 1)) \quad (6)$$

- *C2PSA Block*, a new addition that applies spatial attention to enhance focus on relevant regions, is especially useful for detecting small or occluded objects.

$$C2PSA(X) =$$
$$Attention(Concat(Xpath1, Xpath2)) \quad (7)$$

*2) Neck* aggregates features from different scales and prepares them for detection. It includes [20]:

- *Upsampling and Concatenation Layers* to merge feature maps from various resolutions:

$$Feature\ upsample =$$
$$Upsample(Feature\ previous) \quad (8)$$

$$Feature\ concat =$$
$$Concat(Feature\ upsample, Feature\ lower) \quad (9)$$

- *C3k2 Block* for efficient post-concatenation processing.

$$C3k2\ neck =$$
$$Conv\ small\big(Concat(Feature\ concat)\big) \quad (10)$$

- *C2PSA Block* to apply spatial attention, improving focus in cluttered scenes.

*3) Head:* generates final predictions, including bounding boxes, class labels, and confidence scores. It operates at three scales: P3 (small), P4 (medium), and P5 (large) detection layers, enabling robust detection of objects of varying sizes [20].

$$Detect(P3, P4, P5) = BoundingBoxes + C \quad (11)$$

TABLE IV. YOLO MODELS COMPARISON

| Model | Parameters | | | |
|---|---|---|---|---|
| | mAP@50 | mAP@75 | mAP@[50-95] | FPS |
| YOLOv8 | 73.9% | 64.5% | 45.2% | 260 |
| YOLOv10 | 74.3% | 65.2% | 46.7% | 280 |
| YOLOv11 | 76.8% | 68.1% | 48.5% | 290 |

Table IV presents benchmark analyses comparing YOLOv8, YOLOv10, and YOLOv11 [20]. The results demonstrate that YOLOv11 outperforms the previous models across key evaluation metrics, confirming its superior performance in traffic monitoring tasks.

## II. LITERATURE REVIEW
### A. Related Work

Recent advancements in computer vision and deep learning have significantly influenced the development of intelligent traffic monitoring systems. A wide range of studies have explored the application of YOLO-based models, particularly YOLOv11, for real-time vehicle detection, tracking, and traffic analysis.

Lupian et al. (2025) proposed a YOLOv11-based system for traffic monitoring and accident detection, integrating image processing techniques to enable real-time incident recognition. Their system also supported offline detection via a desktop application, laying the groundwork for localized, infrastructure-independent solutions [12].

Chi (2025) introduced TSM-YOLO, an enhanced YOLOv11 variant optimized for dense traffic and occlusion scenarios. His proposed model, TSM-YOLO, introduces two key improvements: First, the use of the MPDIoU loss function refines the spatial penalty mechanism, thereby improving the precision of bounding box predictions, and second, the integration of SENetV2 channel recalibration enhances the model's discriminative capability for detecting traffic vehicles [2].

Mitra et al. (2025) addressed scalability challenges by leveraging small Uncrewed Aerial Systems (sUAS) to achieve real-time traffic surveillance over large areas. Effectively addressing the scalability challenge by deploying YOLOv11 with StrongSORT and ByteTrack on small uncrewed aerial systems (sUAS), enabling wide-area traffic surveillance from an aerial perspective for vehicle tracking in highway environments. [3].

Tan et al. (2025) developed FF-YOLO, an optimized YOLOv11 framework incorporating three key components: (1) a CA-C3K2 module for robust feature extraction, (2) a spatial-channel mechanism to mitigate occlusion effects, and (3) the MPDIoU loss function for improved localization accuracy. This framework achieved a mean Average Precision (mAP) of 94.2%, representing a +13.7% improvement over YOLOv11n. However, the paper primarily focuses on applications in air traffic control [4].

Dash et al. (2025) demonstrate the superiority of YOLOv11 for vehicle speed estimation, achieving 98.8% accuracy and a 99.2% F1-score on the AI City Challenge dataset, outperforming CNN, LSTM, and EfficientNetB0 models [19]. While this validates YOLOv11's robustness in complex environments, our research extends its application to comprehensive traffic monitoring, focusing on real-time multi-vehicle detection and counting.

Earlier work by Mandal et al. (2020) compared several deep learning models to improve their system, including Mask R-CNN, YOLOv4, and CenterNet, to achieve a real-time vehicle detection system. Among these, the YOLO algorithm achieved the highest performance, with an accuracy of 93.7% in detecting cars, concluding that YOLO-based models outperformed alternatives in terms of accuracy and speed [5].

Zhou et al. (2024) demonstrated the efficacy of YOLOv11 in traffic accident detection, achieving state-of-the-art mean Average Precision (mAP) through architectural innovations such as C3K2, SPPF, and C2PSA, as well as an augmented dataset tailored for complex scenarios [18]. While this validates YOLOv11's potential for safety-critical applications, our research focuses on vehicle detection and counting, with the added capability of storing all detected vehicles for future post-analysis.

Halili et al. (2024) examined the ethical concerns surrounding AI in public spaces in North Macedonia, highlighting issues such as privacy, surveillance, and algorithmic bias. The findings reveal that 47.7% of respondents view AI as a substantial threat to personal privacy, while 61.1% advocate for the establishment of strict regulatory frameworks to govern its use [1].

Alif et al. (2024) analyzed YOLOv11's performance in vehicle detection, emphasizing its architectural improvements over YOLOv8 and YOLOv10. The model demonstrated superior accuracy and robustness, particularly in detecting small and occluded vehicles, while maintaining real-time inference speeds. These enhancements make YOLOv11 highly suitable for intelligent transportation systems and scalable traffic monitoring applications. The study also provides valuable benchmarks that can guide future research in optimizing detection models for complex urban environments [21].

Talaat et al. (2025) proposed a Smart Traffic Management System (STMS) that integrates an enhanced YOLOv11 model for real-time vehicle detection and dynamic traffic flow optimization in urban environments. Their system combines computer vision and AI to analyze live traffic footage, classify vehicle types, and estimate traffic density. Evaluated on real-world datasets, the model achieved a mean Average Precision (mAP) of 92.4% and an Intersection over Union (IoU) of 0.85, demonstrating high detection accuracy and localization precision [21].

Pudaruth et al. (2024) developed a real-time traffic monitoring system using YOLOv8 to address congestion on major roads in Mauritius. Trained on a dataset of 2,800 frames, the system detects, tracks, and counts various vehicle types while estimating traffic density, speed, and bidirectional flow. It achieved high performance, with mean accuracies of 96.1% for counting, 94.4% for classification, and 95.3% for density estimation. Compared to manual methods, the system offers a scalable and cost-effective solution for traffic analysis and congestion mitigation [22].

Borse et al. (2024) proposed a deep learning-based traffic analysis system capable of automatic vehicle detection, tracking, and counting for cars, buses, and trucks. The study employed various YOLO models and conducted a comparative evaluation across multiple performance parameters, with YOLOv8 achieving over 97% accuracy at efficient processing speeds [23].

Saklani et al. (2025) proposed a real-time traffic management system that integrates YOLOv8, Convolutional Neural Networks (CNNs), and Internet of Things (IoT) technologies to enhance urban mobility and reduce congestion. The system dynamically adjusts traffic signals based on real-time vehicle detection and monitoring, leveraging IoT devices for data collection across road networks. A comparative analysis of YOLO versions (v3 to v8) was conducted, with YOLOv8 demonstrating superior performance in terms of detection accuracy, processing speed, and suitability for deployment on edge devices [24].

B. Research Gap and Motivation

Despite the promising results achieved by prior studies, several limitations persist in the current state of AI-based traffic monitoring systems:

- *Limited Integration of Counting Mechanisms*: While many systems focus on detection and tracking, few incorporate robust, real-time vehicle counting mechanisms. This limits their utility for traffic flow analysis and congestion management.

- *Offline and Edge Deployment*: Most existing solutions rely on cloud infrastructure or high-end hardware, making them less suitable for deployment in resource-constrained environments or regions with limited connectivity.

- *Environmental Robustness*: Although some models address occlusion and dense traffic, fewer studies systematically evaluate performance under adverse weather or low-light conditions.

- *User Interface and Usability*: Many systems lack intuitive user interfaces, which prevents adoption by non-technical users.

- *Adaptability to Evolving AI Regulations*. System is designed with adaptability in mind, allowing it to respond to evolving legal frameworks such as the EU AI Act [27], GDPR [28], and other regional data protection laws. This is achieved through: Policy-driven configuration, where data storage duration, anonymization thresholds can be adjusted via configuration files without modifying core code.

This research addresses these gaps by developing AIvision, a YOLOv11-based traffic monitoring system that integrates real-time detection, tracking, and counting within a lightweight desktop application. The system is designed for offline use, supports user-defined detection zones, and includes performance optimizations such as video caching for enhanced speed. These contributions aim to bridge the gap between academic prototypes and deployable smart city solutions.

III. COMPUTER VISION AND IMAGE OBJECT RECOGNITION TECHNOLOGY

According to Barrow, H. G., & Tenenbaum, J. M. (1981), "Vision is an information processing task with well-defined input and output. The input consists of arrays of brightness values, representing projections of three-dimensional scenes recorded by a camera or comparable imaging device. The desired output is a concise description of the three-dimensional scene depicted in the image, the exact nature of which depends upon the goals and expectations of the observer".

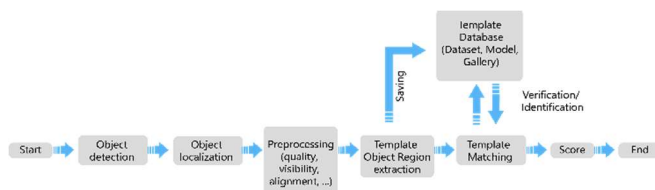

Fig. 6  Processing Phases of an Object Recognition System

Fig.6 illustrates the sequential stages of the object recognition pipeline, from initial detection to final matching,

highlighting the transformation of raw image data into structured semantic information.

In the context of image-based object recognition systems, the process typically involves three primary stages: object detection, template extraction, and template matching [6]

- *Detection Phase*: A frame from the video or image stream is analyzed and scanned to identify the presence of objects. Detected objects are extracted, and if necessary, rescaled and normalized to ensure consistency in further processing.
- *Template Extraction Phase*: The images obtained during detection are processed to classify and filter out low-quality templates. Additional image enhancement

  and normalization may be applied before storing the refined templates in a structured dataset.
- *Template Matching Phase*: The system compares the detected objects with the stored templates to evaluate similarity and determine recognition accuracy.

## IV. VEHICLE VERIFICATION PROCESS

The vehicle recognition process involves the verification of objects detected in recorded images. Fig.7 illustrates the diagram of the vehicle recognition workflow, which comprises two main components: the object registration unit and the object verification unit.

The process within the registration unit begins with the capturing of images from an input device. If necessary, the images undergo pre-processing to enhance their quality, followed by a quality assessment phase. If the image quality meets the required standards, the image is forwarded to the verification unit.

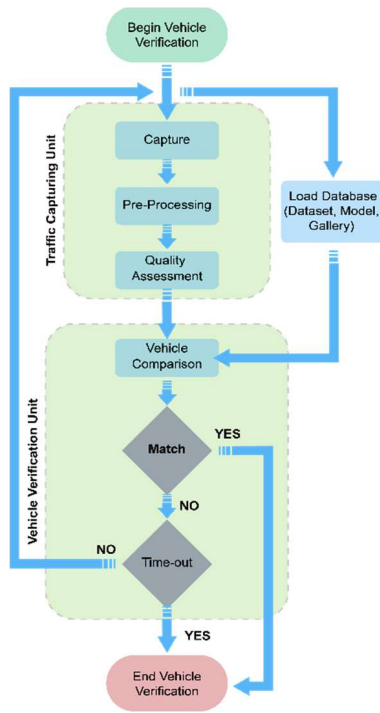

Fig. 7  Vehicle Verification Process

In this unit, the captured image is initially compared against a reference dataset. If a match is found, the verification is deemed successful. Otherwise, the system checks whether the predefined verification time has elapsed. If the time limit has been exceeded, the process restarts; if not, the verification attempt is considered unsuccessful, and the process terminates for that image. Throughout this procedure, newly captured images from the input device are continuously fed into the system. The subjects of interest within these images are compared against templates stored in the dataset. If a subject is successfully identified, the system returns the corresponding match along with the calculated accuracy percentage.

## V. AIVISION VEHICLE DETECTION SYSTEM

According to Halili et al. (2022), classification automation represents a critical and distinctive component within intelligent vision systems.

In light of recent advancements in computer vision, particularly the evolution of YOLO-based AI models, we developed AIvision, an intelligent traffic monitoring system designed to detect, track, and count vehicles on roadways. AIvision leverages the YOLOv11 model, recognized as a state-of-the-art solution in object detection, and is implemented using the PyTorch deep learning framework. For object tracking, the system integrates two advanced algorithms: BoT-SORT and ByteTrack. The model was trained on the COCO dataset [8], which provides a diverse set of annotated images suitable for real-world traffic scenarios.

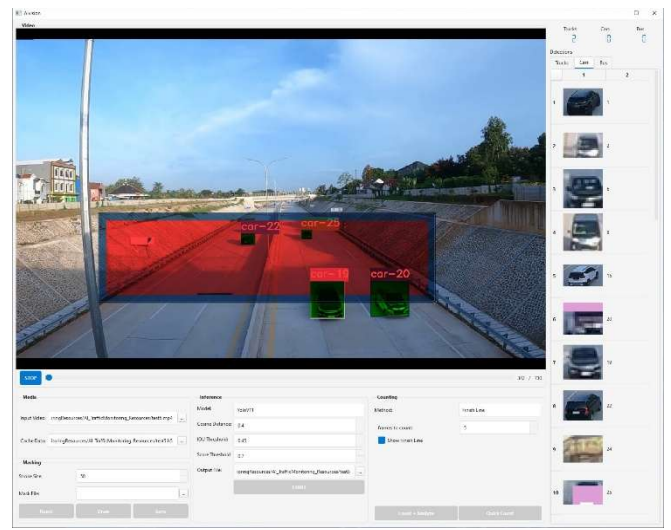

Fig. 8  AIvision Vehicle Detection System Finish Line Method

Given hardware limitations, we implemented two methods for vehicle detection and tracking. The first method processes the video stream directly from the source. The second, more efficient method, involves generating a cached version of the video after initial processing, which improves performance by a factor of 4-5×. This caching mechanism significantly accelerates subsequent analysis.

Vehicle detection in AIvision is performed across all video frames, while tracking is enhanced through a custom-developed method that improves accuracy and robustness. The system also includes a user-friendly interface that allows users to define and adjust detection zones interactively. Additionally, a masking tool enables users to exclude specific regions of the video from the process.

Fig. 8 & Fig. 9 illustrate the AIvision user interface, highlighting all essential components, including detection zones, masking tools, and real-time analytics video frames.

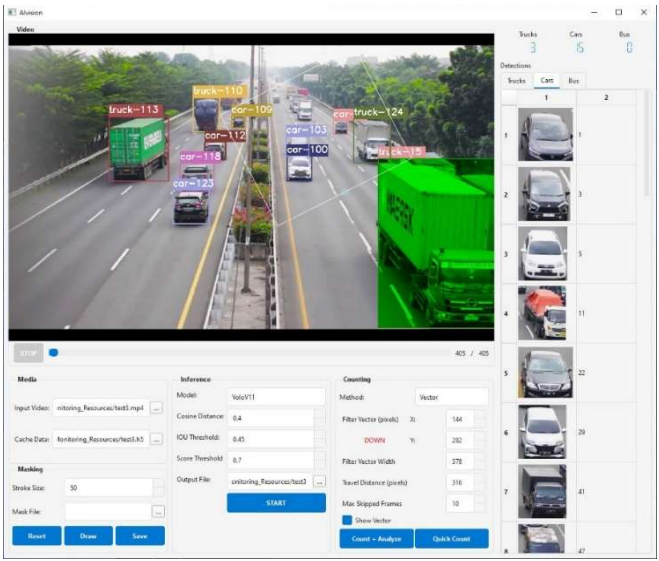

Fig. 9 AIvision Vehicle Detection System Motion Vector Method

Vehicle counting is supported through two distinct methods: Finish Line and Motion Vector. In the Finish Line method, a designated zone is defined within the video frame. A vehicle is counted if it remains within this zone for a specified number of frames set to five by default, but adjustable within the application. The Motion Vector method, on the other hand, is based on three parameters: direction, distance, and width. These parameters define the motion path and are particularly effective in scenarios where the direction of vehicle movement is critical for traffic analysis.

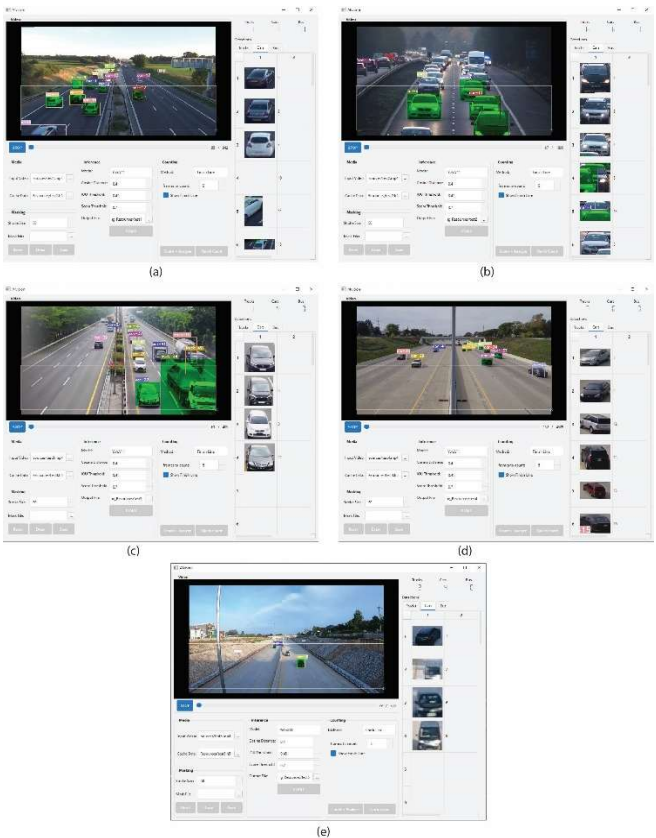

Fig. 10 Tested video Aivision UI (a) test1.mp4 - February, 03:43 pm, (b) test2.mp4 - September, 05:51 pm, (c) test3.mp4 - January, 08:14 am, (d) test4.mp4 - November, 10:42 am, (e) test5.mp4 - June, 04:14 pm.

Fig.10 shows the AIvision performing detection and counting in real time. All five video frames were used. The test conducted on this paper uses the following parameters for the YOLOv11 model: Cosine Distance 0.4, IOU threshold 0.45, Score threshold 0.7, main counting method Finish Line, and Frames to count 5.

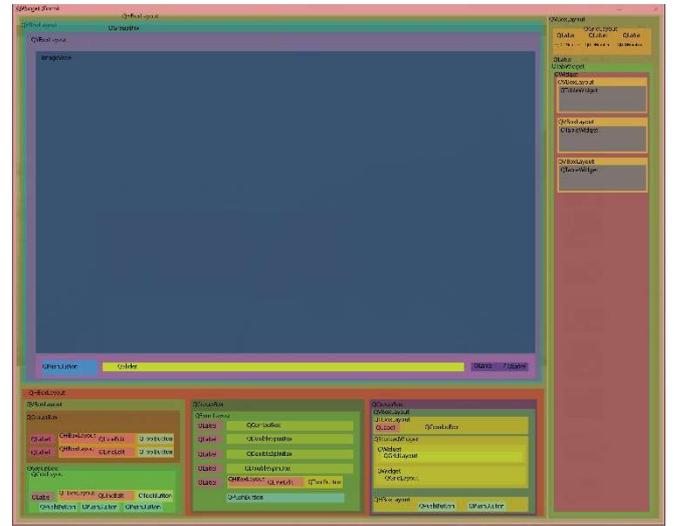

Fig. 11 Qt Component Structure of the System UI

The graphical user interface (GUI) of the AIvision system is developed using the Qt framework for Python, providing a modular, responsive, and user-friendly environment for traffic monitoring operations. The interface adheres to Material Design principles, ensuring clarity, consistency, and ease of use across all components. The application layout is structured into two primary columns:

1) *The left column:*

   a) *Video Display Panel:* Positioned at the top, this panel renders the input or cached video stream in real time.

   b) *Control Panel:* Located beneath the video display, it includes input fields, dropdown menus, and action buttons for managing system operations such as video selection, mask application, model configuration, and execution.

2) *The right column:*

   a) *Masking Tool:* Displays real-time vehicle counting metrics, including category-wise counts (e.g., cars, trucks, buses).

   b) *Detection Gallery:* Shows thumbnails of detected vehicles grouped by class, enabling visual verification and post-analysis.

3) *Additional UI components include:*

   a) *Analytics Panel:* Allows users to define regions of interest (ROI) by drawing exclusion zones directly on the video frame. Saved masks are automatically loaded in subsequent sessions.

   b) *Counting Configuration Module*: Enables selection between Finish Line and Motion Vector methods, with adjustable parameters for frame thresholds, direction vectors, and zone dimensions.

   c) *Status Console*: Provides feedback on system operations, including detection progress, frame rate, and error messages.

*d) File Management Dialogs*: Facilitate input/output path selection for video files, cached data, and stored vehicle images.

This component-based architecture ensures that users can intuitively navigate the system, configure detection and analysis parameters, and visualize results in real time. The modular design also supports future extensibility, allowing additional features to be integrated with minimal disruption to the existing interface.

The UML sequential diagram below outlines the operational flow of a traffic monitoring system that integrates the YOLOv11 object detection model. The process follows these steps:

*4) Region of Interest Masking (Optional):* The user may choose to apply a mask to define a region of interest. If a mask file exists, it is loaded. Otherwise, the user can draw and save a new mask file.

*5) Model Configuration and Execution:* The user configures the model parameters and defines the output path. Upon clicking "Start", the system executes the YOLOv11 model. Performs object detection and generates a cached video with annotated results.

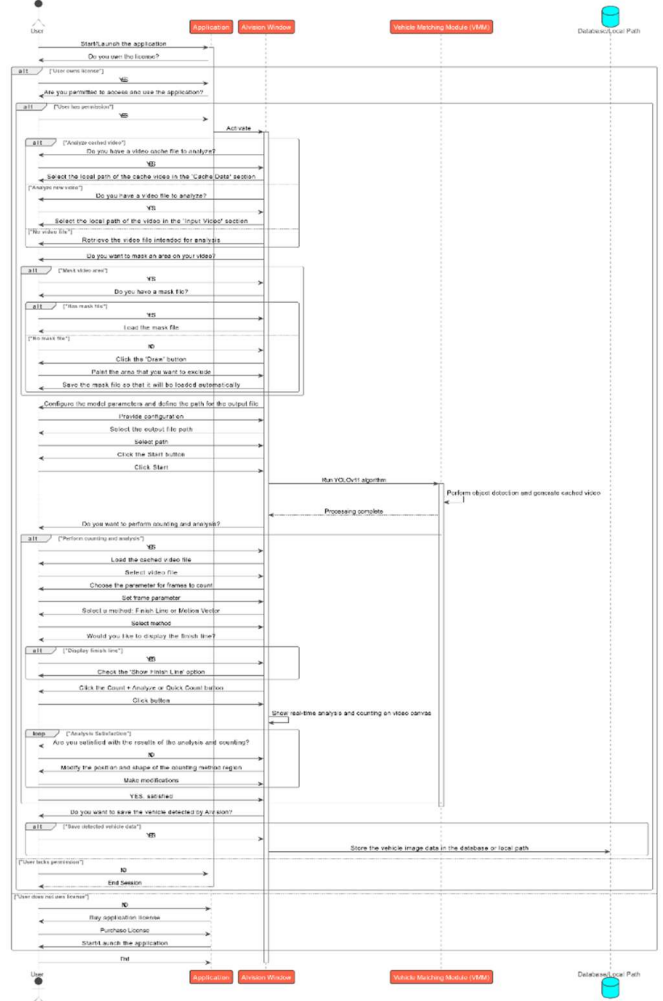

Fig. 12  Sequential Diagram of the traffic monitoring system

*6) Vehicle Counting and Analysis:* If counting is enabled, the cached video is loaded. Frame count parameters are set. The user selects a counting method: Finish Line or Motion Vector. Optionally, the finish line can be displayed for visual reference. The user initiates analysis via "Count + Analyze" or "Quick Count". Real-time analysis results are displayed. The user may iteratively refine the counting region or method until satisfied with the results.

*7) Data Storage (Optional):* If enabled, detected vehicle data is stored either in a local database or on the device.

*8) Session Termination:* The session concludes when the user ends the application.

Fig.12 & Fig.13 illustrate a sequential diagram and flowchart that models the operational workflow of the AIVision vehicle detection and analysis system.

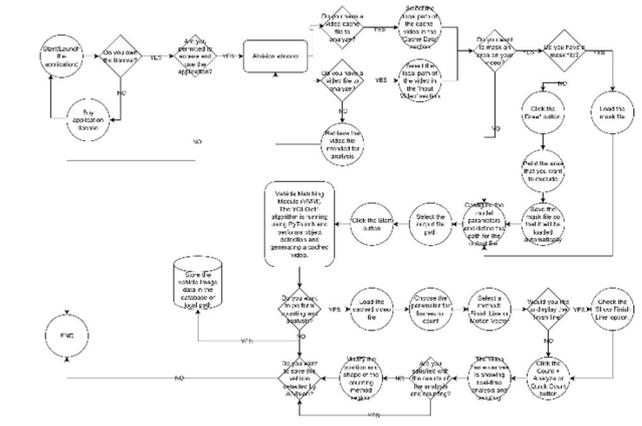

Fig. 13  Flowchart diagram

## VI. SYSTEM METRICS

Following the development of the Alpha version of the traffic monitoring system, AIvision, a series of tests were conducted. The results are summarized in Tables V & VI. The evaluation involved five randomly selected video recordings, captured at different locations, under varying atmospheric conditions, and at different times of day. Although the videos varied in duration, file size, and frame count, they all shared a common resolution of 1280×720 pixels. The object categories targeted for detection included cars, buses, and trucks.

TABLE V.  TRAFFIC MONITORING ENGINE PERFORMANCE METRICS

| (1) | (2) | (3) | (4) | (5) | (6) | (7) | (8) | (9) | (10) | (11) |
|---|---|---|---|---|---|---|---|---|---|---|
| 1 | 942 | 31 | 126 | 30 | (a) | 61 | 63 | 63 | 103.28 | 103.28 |
|  |  |  |  |  | (b) | 0 | 0 | 0 | 0.0 | 0.0 |
|  |  |  |  |  | (c) | 13 | 11 | 11 | 84.62 | 84.62 |
| 2 | 1,800 | 60 | 312 | 57 | (a) | 48 | 46 | 46 | 95.83 | 95.83 |
|  |  |  |  |  | (b) | 0 | 0 | 0 | 0.0 | 0.0 |
|  |  |  |  |  | (c) | 7 | 4 | 4 | 57.14 | 57.14 |
| 3 | 405 | 16 | 53 | 12 | (a) | 30 | 26 | 26 | 86.67 | 86.67 |
|  |  |  |  |  | (b) | 0 | 0 | 0 | 0.0 | 0.0 |
|  |  |  |  |  | (c) | 21 | 14 | 14 | 66.67 | 66.67 |
| 4 | 4,509 | 150 | 502 | 144 | (a) | 311 | 294 | 294 | 94.53 | 94.53 |
|  |  |  |  |  | (b) | 0 | 0 | 0 | 0.0 | 0.00 |
|  |  |  |  |  | (c) | 6 | 4 | 4 | 66.67 | 66.67 |
| 5 | 730 | 30 | 103 | 23 | (a) | 14 | 11 | 11 | 78.57 | 78.57 |
|  |  |  |  |  | (b) | 0 | 0 | 0 | 0.0 | 0.00 |
|  |  |  |  |  | (c) | 2 | 2 | 2 | 100.0 | 100.0 |

(1)-Video; (2)- Nr. Frames; (3)- Video Duration; (4)- Video Processing Duration; (5)- Cached Video Processing Duration; (6)- Vehicle type; (7)- Ground Truth Nr. of Vehicles; (8)- Total Number of Detections from video; (9)- Total Number of Detections from cache; (10)- Video Precision / Counting accuracy (%); (11)- Cache Video Precision / Counting accuracy (%).

**Measurable units**: Frames - fs; **Video Duration** - sec; **Video Resolution** - pixel; **Video Size** - kb; **Ground Truth Total Number**,

| Total Number of Detections - unit(n); Precision, Counting accuracy - percent (%) (a)-Car; (b)-Bus, (c)-Truck |
|---|
| **Computer Specifications**: **CPU**: Intel(R) Core(TM) i7-6700HQ CPU @ 2.60GHz; **GPU**: NVidia GTX 960M 4.0 GB; **RAM**: 32.0 GB; **OS**: Windows 10 Pro 22H2; **Project Code URL alpha-v1**: https://github.com/shkelqimsherifi/YOLOv11_TrafficMonitoring; Video Links [9]. |

Table V. presents a comparative analysis of detection speed and the number of objects identified in both the original and cached versions of the videos.

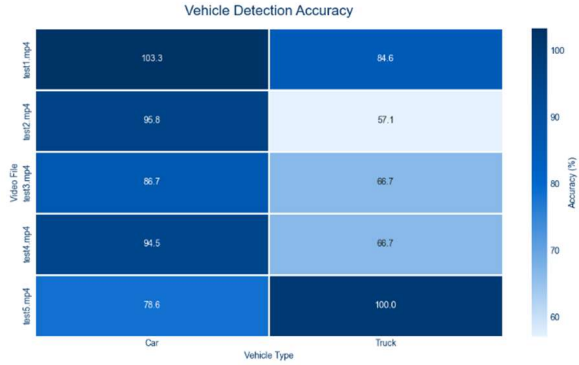

Fig. 14  Vehicle Detection Accuracy Comparison

Fig.14 presents the performance of the YOLOv11 model across five test videos, focusing on two vehicle categories: cars and trucks. Detection accuracy varied from 57.1% to 103.3%, with the latter indicating a likely labeling or post-processing anomaly. Overall, car detection tended to be more consistent and generally higher than truck detection, which showed greater variability across videos. For instance, truck detection accuracy dropped as low as 57.1% in test2.mp4, while car detection remained above 78% in all cases. These results suggest that the model performs more reliably on smaller, more frequently occurring vehicle types, while larger or less common classes like trucks may require further tuning or dataset balancing.

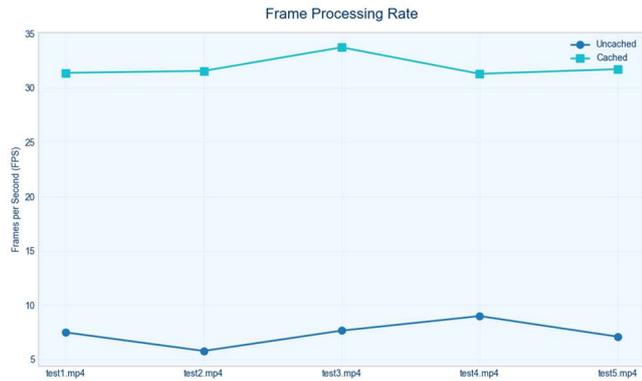

Fig. 15  Frame Processing Rate

Fig.15 compares the YOLOv11 system's performance on five test videos. It shows that uncached processing runs at 5-10 FPS, while cached processing maintains a consistent 30-35 FPS. This highlights the significant speed advantage of using cached data for near real-time analysis.

Fig.16 compares the number of vehicles detected by the YOLOv11 model against manually annotated ground truth data across five test videos (test1.mp4 to test5.mp4). The x-axis lists the video files, while the y-axis shows vehicle counts ranging from 0 to 300. For each video, light blue bars represent detected trucks, dark blue bars represent detected cars, and dark blue dots indicate the total ground truth vehicle count. This visualization highlights the model's detection accuracy per vehicle type and its alignment with the annotated reference data.

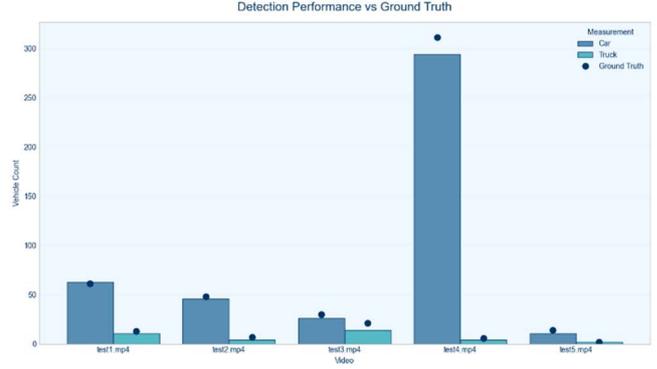

Fig. 16  Detection Performance

Fig.17 compares the processing times of five test videos using uncached and cached modes within the YOLOv11-based traffic monitoring system. The y-axis represents processing time in seconds (0-500), while the x-axis lists the video files (test1.mp4 to test5.mp4). Each bar is segmented into uncached (dark blue) and Cached (light blue) portions, with efficiency multipliers annotated above. Cached processing significantly reduces execution time, achieving speedups ranging from 3.5× to 5.5×, demonstrating the system's improved efficiency when reusing pre-processed data.

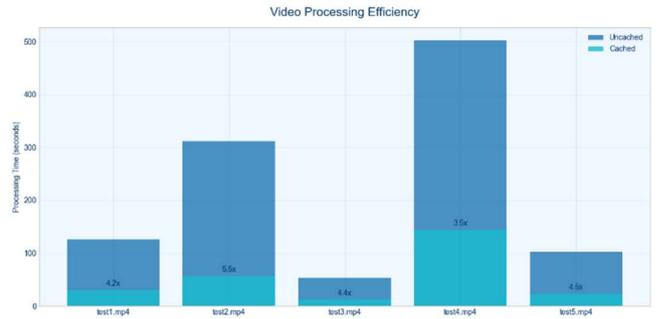

Fig. 17  Video Processing Efficiency

TABLE VI.  TRAFFIC MONITORING ENGINE PERFORMANCE METRICS

| (1) | (2) | (3) | (4) | (5) | (6) | (7) | (8) | (9) |
|---|---|---|---|---|---|---|---|---|
| 1 | (a) | 61 | 63 | 2 | 0 | 0.032 | 0 | 0.5 |
|   | (b) | 0 | 0 | 0 | 0 | 0 | 0 | 0.5 |
|   | (c) | 13 | 11 | 0 | 2 | 0 | 0.153 | 0.5 |
| 2 | (a) | 48 | 46 | 0 | 2 | 0.04 | 0 | 0.5 |
|   | (b) | 0 | 0 | 0 | 0 | 0 | 0 | 0.5 |
|   | (c) | 7 | 4 | 0 | 3 | 0 | 0.43 | 0.5 |
| 3 | (a) | 30 | 26 | 0 | 4 | 0 | 0.13 | 0.5 |
|   | (b) | 0 | 0 | 0 | 0 | 0 | 0 | 0.5 |
|   | (c) | 21 | 14 | 0 | 7 | 0 | 0.33 | 0.5 |
| 4 | (a) | 311 | 294 | 0 | 17 | 0 | 0.054 | 0.5 |
|   | (b) | 0 | 0 | 0 | 0 | 0 | 0 | 0.5 |
|   | (c) | 6 | 4 | 0 | 2 | 0 | 0.33 | 0.5 |
| 5 | (a) | 14 | 11 | 0 | 3 | 0 | 0.21 | 0.5 |
|   | (b) | 0 | 0 | 0 | 0 | 0 | 0 | 0.5 |
|   | (c) | 2 | 2 | 0 | 0 | 0 | 0 | 0.5 |
| (1)-Video; (2)- Vehicle type; (3)- Ground Truth Nr. of Vehicles; (4)- Total Number of Detections from video; (5)- False Positive; (6)- False Negative; (7)- FPR- False Positive Ratio; (8)- False Negative Ratio; (9)- IoU ||||||||||
| **Type of vehicle**: (a)-Car; (b)-Bus, (c)-Truck ||||||||||
| **Computer Specifications**: **CPU**: Intel(R) Core(TM) i7-6700HQ CPU @ 2.60GHz; **GPU**: NVidia GTX 960M 4.0 GB; **RAM**: 32.0 GB; **OS**: Windows 10 Pro 22H2; ||||||||||

| **Project Code URL alpha-v1**: |
| --- |
| https://github.com/shkelqimsherifi/YOLOv11_TrafficMonitoring; Video Links [9] |

Fig.18 presents a comparative analysis of the False Positive Rate (FPR) and False Negative Rate (FNR) across five video samples used in the evaluation of the YOLOv11-based traffic monitoring system. False Positive Rate (FPR) ranging from 0.00 to 0.032, suggesting high detection reliability under its specific conditions. Test1.mp4 shows the highest error rates among all samples, 0.032, indicating challenges such as complex traffic scenes and not optimal lighting. False Negative Rate (FNR) ranging from 0.00 to 0.43, suggesting high detection reliability under its specific conditions. Test2.mp4 shows the highest error rates among all samples, 0.43, indicating challenges such as complex traffic scenes and not optimal lighting.

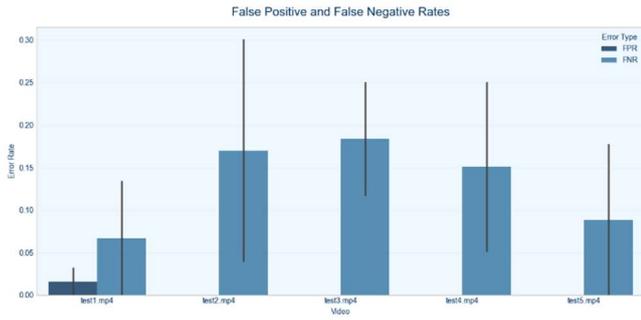

Fig. 18  Error Rate

TABLE VII.  Traffic Monitoring Engine Performance Metrics

| (1) | (2) | (3) | (4) | (5) | (6) | (7) | (8) | (9) |
| --- | --- | --- | --- | --- | --- | --- | --- | --- |
| 1 | (a) | 61 | 63 | 2 | 0 | 0.97 | 1 | 0.98 |
|   | (b) | 0 | 0 | 0 | 0 |   |   |   |
|   | (c) | 13 | 11 | 0 | 2 | 1 | 0.86 | 0.93 |
| 2 | (a) | 48 | 46 | 0 | 2 | 1 | 0.96 | 0.98 |
|   | (b) | 0 | 0 | 0 | 0 |   |   |   |
|   | (c) | 7 | 4 | 0 | 3 | 1 | 0.7 | 0.82 |
| 3 | (a) | 30 | 26 | 0 | 4 | 1 | 0.88 | 0.94 |
|   | (b) | 0 | 0 | 0 | 0 |   |   |   |
|   | (c) | 21 | 14 | 0 | 7 | 1 | 0.75 | 0.86 |
| 4 | (a) | 311 | 294 | 0 | 17 | 1 | 0.95 | 0.97 |
|   | (b) | 0 | 0 | 0 | 0 |   |   |   |
|   | (c) | 6 | 4 | 0 | 2 | 1 | 0.75 | 0.86 |
| 5 | (a) | 14 | 11 | 0 | 3 | 1 | 0.82 | 0.9 |
|   | (b) | 0 | 0 | 0 | 0 |   |   |   |
|   | (c) | 2 | 2 | 0 | 0 | 1 | 1 | 1 |
| (1)-Video; (2)- Vehicle type; (3)- Ground Truth Nr. of Vehicles; (4)- Total Number of Detections from video; (5)- False Positive; (6)- False Negative; (7)- Precision; (8) - Recall; (9)- F1 Score ||||||||| 
| **Type of vehicle**: (a)-Car; (b)-Bus, (c)-Truck ||||||||| 
| **Computer Specifications**: **CPU**: Intel(R) Core(TM) i7-6700HQ CPU @ 2.60GHz; **GPU**: NVidia GTX 960M 4.0 GB; **RAM**: 32.0 GB; **OS**: Windows 10 Pro 22H2; **Project Code URL alpha-v1**: https://github.com/shkelqimsherifi/YOLOv11_TrafficMonitoring; Video Links [9] |||||||||

In evaluating the performance of the YOLOv11 model for traffic monitoring, precision metrics were analyzed across multiple video datasets and vehicle categories. The results, in Fig. 19, demonstrate consistently high precision values, indicating the model's robustness in distinguishing between vehicle types under varying conditions. Specifically, the model achieved a precision of 0.968 for cars in test1.mp4, while attaining perfect precision 1.0 for trucks in test2.mp4 and test5.mp4, as well as for cars in test4.mp4. These findings suggest that YOLOv11 is highly effective in minimizing false positives, particularly for truck detection, where precision remained consistently at the maximum value.

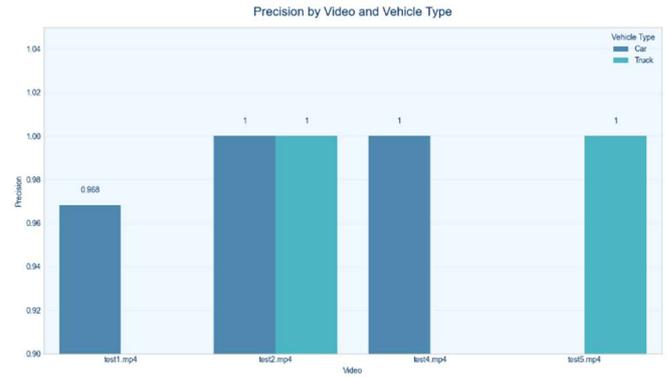

Fig. 19  Precision

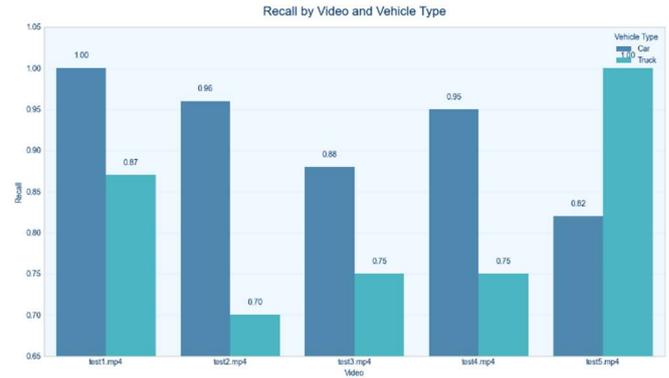

Fig. 20  Recall

The recall performance of the YOLOv11 model was evaluated across five video sequences, with results disaggregated by vehicle type, cars, and trucks. The model demonstrated high recall for cars, achieving 1.00 in test1.mp4, 0.96 in test2.mp4, 0.88 in test3.mp4, 0.95 in test4.mp4, and 0.82 in test5.mp4. In contrast, the recall for trucks showed greater variability, with values of 1.00 in test5.mp4, 0.86 in test1.mp4, 0.7 in test3.mp4, 0.75 in test4.mp4, and a lower score of 0.70 in test2.mp4. These results indicate that while the YOLOv11 model maintains strong recall for car detection across diverse video contexts, its performance in detecting trucks is less consistent. This discrepancy may be attributed to factors such as fewer training samples for trucks, occlusions, or differences in vehicle scale and motion dynamics. The findings underscore the need for targeted improvements in truck detection to ensure balanced recall across all vehicle categories.

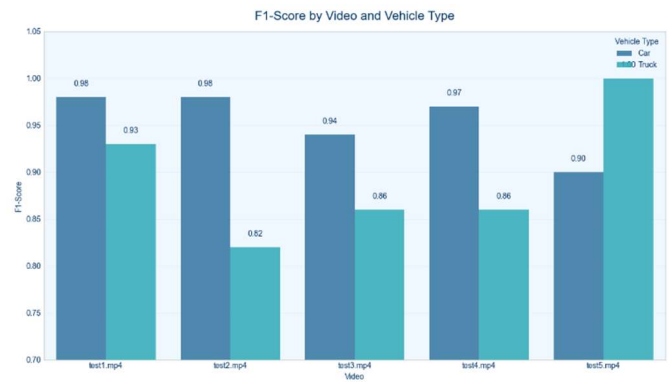

Fig. 21  F1-Score

The F1-score, which harmonizes precision and recall into a single metric, was used to evaluate the overall detection performance of the YOLOv11 model across five video datasets for two vehicle categories: cars and trucks. As shown in Fig. 21, the model consistently achieved high F1-scores for cars, with values of 0.98 in test1.mp4 and test2.mp4, 0.97 in test4.mp4, 0.94 in test3.mp4, and 0.90 in test5.mp4. In contrast, the F1-scores for trucks range from 0.93 in test1.mp4 to 0.82 in test2.mp4, 0.86 in test3.mp4 and test4.mp4, and peaking at 1.00 in test5.mp4. These results suggest that while the YOLOv11 model maintains robust and consistent performance for car detection, truck detection is more sensitive to contextual factors such as video quality, lighting conditions, and occlusion.

## VII. CONCLUSION

This study presents the development and evaluation of AIvision, an AI-based traffic monitoring system designed to operate in real time with high accuracy. Leveraging the YOLOv11 model and integrated tracking algorithms, the system demonstrates strong performance in vehicle detection, tracking, and counting across diverse environmental conditions.

The YOLOv11-based traffic monitoring system, AIvision, demonstrated high detection accuracy ranging from 66.67% to 95.83% across various test scenarios. Initial video processing was observed to be 3.5 to 5.5 times slower than real-time, primarily due to computational overhead during the first pass. However, the system subsequently generates a cached version of the processed video, enabling real-time performance on all tested sequences. Precision scores ranged from 0.97 to 1.00 for cars and 1.00 for trucks, indicating minimal false positives. Recall values were 0.82-1.00 for cars and 0.70-1.00 for trucks, suggesting stronger performance for cars, likely due to a larger volume of training data and more consistent visual features. F1-scores further confirmed this trend, with 0.90-1.00 for cars and 0.82-1.00 for trucks, highlighting the model's robustness in car detection and slightly more variability in truck detection.

Despite these promising results, the system's performance degrades under adverse weather conditions, and it has not yet been tested or deployed in large-scale multi-camera environments. Nevertheless, the ability to generate cached video enables the possibility of performing initial computation in the cloud, followed by real-time playback at approximately 30 FPS on mid-range GPUs. This hybrid approach offers a practical pathway for scaling the system while maintaining responsiveness.

While adverse weather conditions do impact detection performance, the system remains functional, underscoring the robustness of the underlying model and implementation.

## VIII. DISCUSSION

### A. Future Work and Limitations

A key direction for future work involves a comprehensive performance evaluation under diverse operational conditions. This includes measuring processing speed (frames per second, FPS) across different hardware configurations (e.g., CPU-only, GPU-accelerated, edge devices), as well as conducting a detailed memory usage analysis and processing latency profiling. These metrics are essential for optimizing the system for deployment in resource-constrained environments.

Additionally, a comparative analysis should be conducted to benchmark the proposed system against:

- Traditional traffic monitoring methods, such as loop detectors and manual counting.
- Other AI-based approaches, including alternative object detection models (e.g., Faster R-CNN, EfficientDet),
- Commercial traffic monitoring solutions, to evaluate cost-effectiveness, scalability, and deployment feasibility.

The current study also acknowledges several limitations:

- *First*, scalability remains a concern, particularly when extending the system to city-wide deployments involving multiple camera feeds and high traffic density.
- *Second*, while the system achieves near real-time performance, further optimization of inference speed and tracking efficiency is necessary for high-throughput environments.
- *Third*, integration with existing traffic infrastructure, such as traffic signal control systems and centralized traffic management platforms, has not yet been implemented or tested.
- *Finally*, privacy considerations must be addressed, especially in public surveillance contexts, by incorporating techniques such as anonymization, edge processing, and compliance with data protection regulations.

Addressing these limitations will be critical for transitioning from a prototype to a production-ready system capable of supporting intelligent transportation systems and smart city initiatives.

TABLE V.     NOMENCLATURE

| Acronyms and abbreviations | Parameters |
|---|---|
| FPS | Frames Per Second |
| AI | Artificial Intelligence |
| FN | False Negative |
| FP | False Positive |
| GT | Ground Truth |
| IoU | Intersection over Union |
| MOT | Multi-Object Tracking |
| MOTA | Multi-Object Tracking Accuracy |
| MOTP | Multi-Object Tracking Precision |
| TP | True Positive |
| YOLO | You Only Look Once |
| IDS | Identity Switch |

## REFERENCES

[1] Bajrami, E., & Halili, F. (2024). Exploring The Impact Of Artificial Intelligence On The IoT And Digital Agenda In The Western Balkans: Integrating A Proposed Web Application For Regional Advancement. *JNSM Journal of Natural Sciences and Mathematics of UT, 9(17-18), 244-258.*

[2] Chi, B. (2025, March). TSM-YOLO: An Optimized YOLOv11 Model for Traffic State Surveillance. *In 2025 7th International Conference on Software Engineering and Computer Science (CSECS) (pp. 1-6). IEEE.*

[3] Mitra, R., Sourav, M. A. A., Kim, S., Gulmezoglu, B., & Ceylan, H. (2025). Comparative Case Study: Traffic Monitoring Using YOLOv11-Based Object Detection and Two Tracking Algorithms with Small Uncrewed Aerial Systems. *In International Conference on Transportation and Development 2025 (pp. 311-321).*

[4] Tan, S., Pan, W., Deng, L., Zuo, Q., & Zheng, Y. (2025). FF-YOLO: An Improved YOLO11-Based Fatigue Detection Algorithm for Air Traffic Controllers. *Applied Sciences, 15(13), 7503*.

[5] Mandal, V., Mussah, A. R., Jin, P., & Adu-Gyamfi, Y. (2020). *Artificial intelligence-enabled traffic monitoring system. Sustainability, 12(21), 9177*.

[6] Majumder, D. D. (1988). Computer Vision and Knowledge-Based Computer Systems. *IETE Journal of Research, 34(3), 230-245*.

[7] Halili, F., Alihajdaraj, E., Berisha, N. & Shoshi, L. (2019). *Artificial Intelligence In Work Process Automation*.

[8] COCO, 2025. [Online]. Available: https://cocodataset.org/#home, (Accessed: 10 April 2025).

[9] Test Videos, 2025. [Online]. Available: https://drive.google.com/drive/folders/1aZZqMP9EWoQ-XYmMbMDHQeUv7l8_Z_5G?usp=sharing

[10] Fetzer, J. H. (1990). What is artificial intelligence?. In Artificial Intelligence: Its scope and limits (pp. 3-27). *Dordrecht: Springer Netherlands*.

[11] Gates, E. (1989). Webster's New World Dictionary. *English Today, 5(2), 52-54*.

[12] Lupian, R. R. F., Arong, C. G., Betinol, W. S., & Valdez, D. B. (2025, April). Intelligent Traffic Monitoring And Accident Detection System Using YOLOv11 And Image Processing. *In 2025 IEEE Open Conference of Electrical, Electronic, and Information Sciences (eStream) (pp. 1-5)*.

[13] Spahija, L., & Halili, F. (2017). Review of artificial intelligence development, its impact, and its challenges.

[14] Barrow, H. G., & Tenenbaum, J. M. (1981). Computational vision. *Proceedings of the IEEE, 69(5), 572-595*.

[15] Festim Halili, Avni Rustemi, Predictive Modeling: Data Mining Regression Technique Applied in a Prototype. *In International Journal of Computer Science and Mobile Computing, vol. 5, issue. 8, 2015*.

[16] Bochkovskiy, A., Wang, C. Y., & Liao, H. Y. M. (2020). Yolov4: Optimal speed and accuracy of object detection. *arXiv preprint arXiv:2004.10934*.

[17] Bakirci, M., Dmytrovych, P., Bayraktar, I., & Anatoliyovych, O. (2024, October). Multi-class vehicle detection and classification with YOLO11 on UAV-captured aerial imagery. *In 2024 IEEE 7th International Conference on Actual Problems of Unmanned Aerial Vehicles Development (APUAVD) (pp. 191-196). IEEE*.

[18] Zhou, Z. (2024, December). Traffic accident detection based on YOLOv11. *In 2024, IEEE 2nd International Conference on Electrical, Automation and Computer Engineering (ICEACE) (pp. 363-369). IEEE*.

[19] Dash, S., Padhy, S., Das, S. K., & Mishra, D. (2025, May). A Comparative Deep Learning Approach for Vehicle Speed Monitoring using YoLov11. *In 2025 International Conference on Intelligent and Cloud Computing (ICoICC) (pp. 1-6). IEEE*.

[20] Alif, M. A. R. (2024). Yolov11 for vehicle detection: Advancements, performance, and applications in intelligent transportation systems. *arXiv preprint arXiv:2410.22898*.

[21] Talaat, F. M., El-Balka, R. M., Sweidan, S., Gamel, S. A., & Al-Zoghby, A. M. (2025). Smart traffic management system using YOLOv11 for real-time vehicle detection and dynamic flow optimization in smart cities. *Neural Computing and Applications, 1-18*.

[22] Pudaruth, S., & Boodhun, I. M. (2024). Reducing Traffic Congestion Using Real-Time Traffic Monitoring with YOLOv8. *International Journal of Advanced Computer Science & Applications, 15(10)*.

[23] Borse, R., Bhattacharyya, A., Sarkar, A., & Bhattacharjee, S. (2024, November). Employing the YOLO model for traffic monitoring on roadways. *In the 2024 International Conference on Intelligent Computing and Sustainable Innovations in Technology (IC-SIT) (pp. 1-6). IEEE*.

[24] Saklani, S., Dhondiyal, S. A., & Singh, D. (2025, April). Real-Time Traffic Management System Using YOLOv8: An analysis of various YOLO Models. *In 2025 4th OPJU International Technology Conference (OTCON) on Smart Computing for Innovation and Advancement in Industry 5.0 (pp. 1-6). IEEE*.

[25] Lin, T. Y., Maire, M., Belongie, S., Hays, J., Perona, P., Ramanan, D., & Zitnick, C. L. (2014, September). Microsoft Coco: Common objects in context. *In European Conference on Computer Vision (pp. 740-755). Cham: Springer International Publishing*.

[26] Milan, A., Leal-Taixé, L., Reid, I., Roth, S., & Schindler, K. (2016). MOT16: A benchmark for multi-object tracking. *arXiv preprint arXiv:1603.00831*.

[27] Regulation, P. (2016). Regulation (EU) 2016/679 of the European Parliament and of the Council. *Regulation (EU), 679(2016), 10-13*.

[28] The impact of the General Data Protection Regulation (GDPR) on artificial intelligence, 2020. [Online]. Available: https://www.europarl.europa.eu/RegData/etudes/STUD/2020/641530/EPRS_STU(2020)641530_EN.pdf, (Accessed: 24 July 2025).